\documentclass[conference]{IEEEtran}
\IEEEoverridecommandlockouts

\usepackage{cite}
\usepackage{amsmath,amssymb,amsfonts}
\usepackage{graphicx}
\usepackage{textcomp}
\usepackage{xcolor}
\usepackage{booktabs}
\usepackage{multirow}
\usepackage{url}
\usepackage{microtype}

\def\BibTeX{{\rm B\kern-.05em{\sc i\kern-.025em b}\kern-.08em
    T\kern-.1667em\lower.7ex\hbox{E}\kern-.125emX}}

\begin{document}

\title{RelCheck: Dual-Evidence Spatial Grounding\\
for VLM Hallucination Correction}

\author{
\IEEEauthorblockN{Siddhi Patil}
\IEEEauthorblockA{\textit{Dept. of Computer Science}\\
\textit{San Jos\'{e} State University}\\
siddhi.patil@sjsu.edu}
\and
\IEEEauthorblockN{Dr.~Navrati Saxena}
\IEEEauthorblockA{\textit{Dept. of Computer Science}\\
\textit{San Jos\'{e} State University}\\
navrati.saxena@sjsu.edu}
\and
\IEEEauthorblockN{Dr.~William B. Andreopoulos}
\IEEEauthorblockA{\textit{Dept. of Computer Science}\\
\textit{San Jos\'{e} State University}\\
william.andreopoulos@sjsu.edu}
}

\maketitle

\begin{abstract}
Multimodal large language models (MLLMs) frequently generate text that is inconsistent with the input image. While object- and attribute-level hallucinations have received considerable attention, relational hallucinations (incorrect descriptions of spatial or interactive relationships between objects) remain largely unaddressed by existing post-hoc correction methods. We present \textbf{RelCheck}, a training-free post-hoc correction pipeline that augments object-level visual grounding with dual relational evidence: learned scene-graph triples from RelTR and deterministic spatial predicates from bounding-box geometry. These combine with a Woodpecker-style object claim layer to form a three-layer visual knowledge base, which a language model corrector uses to rewrite hallucinated text. Evaluated on LLaVA~v1~13B, RelCheck achieves a total MME hallucination score of 630.0 versus 585.0 for a Woodpecker-style baseline, with the largest gain on the position subtask (+31.7 points, accuracy+ improving from 0.367 to 0.600). A four-configuration ablation confirms that both relational layers contribute independently (McNemar $p=0.025$). These results show that structured relational evidence meaningfully improves post-hoc hallucination correction on the spatial reasoning subtasks where current MLLMs are most deficient.
\end{abstract}


\section{Introduction}

Multimodal large language models (MLLMs) such as LLaVA~\cite{liu2023llava}, MiniGPT-4~\cite{zhu2024minigpt4}, and mPLUG-Owl~\cite{ye2023mplugowl} integrate a pretrained vision encoder with a language model backbone to enable joint reasoning over images and text. Despite strong performance on visual question answering and image captioning, these models frequently produce \textit{hallucinations}: textual outputs that are inconsistent with the input image~\cite{li2023pope,yin2024woodpecker}.

Hallucinations in MLLMs occur at three granularities. \textit{Object-level hallucinations} involve fabricated or omitted objects. \textit{Attribute-level hallucinations} involve incorrect descriptions of object properties. \textit{Relational hallucinations} involve incorrect descriptions of the spatial, positional, or interactive relationships between objects (for example, stating ``the cat is on the table'' when the cat is beneath it). While object- and attribute-level errors have received considerable attention, relational hallucinations remain largely unaddressed by existing correction methods.

Two broad approaches have been proposed to mitigate hallucinations. \textit{Instruction tuning} fine-tunes the MLLM on curated data~\cite{liu2023mitigating}, but requires access to model internals and carries high retraining cost. \textit{Post-hoc correction} operates on the MLLM's output after generation, consulting external visual perception models to detect and rewrite hallucinated content. Woodpecker~\cite{yin2024woodpecker}, the leading post-hoc method, constructs a visual knowledge base from object detections and attribute-level visual question answering (VQA), then uses a language model to rewrite hallucinated text. This training-free design is effective for object and attribute errors, but it does not detect spatial or interactional relationships between objects.

The gap is consequential. Empirical benchmarks show that MLLMs score 20--30 percentage points lower on spatial reasoning queries than on object detection queries in the same images~\cite{kamath2023whatsup}. On the MME benchmark~\cite{fu2023mme}, the position subtask consistently yields the lowest scores across MLLM families. LLaVA~v1~13B achieves an accuracy+ of 0.000 on this subtask, meaning it cannot simultaneously answer both the positive and negative variants of a positional question for any image in the dataset.

We present \textbf{RelCheck}, a training-free post-hoc correction pipeline that addresses this gap by augmenting the object-level knowledge base of Woodpecker with two relational evidence layers: (1) learned subject--predicate--object triples from RelTR~\cite{cong2023reltr}, a transformer-based scene graph generator, and (2) deterministic spatial predicates computed from GroundingDINO~\cite{liu2023groundingdino} bounding box coordinates. Together these form a three-layer visual knowledge base that captures the spatial and interactional structure of a scene in ways that prior methods cannot.

Consider an image where a cat is wearing a hat on its head. LLaVA~v1 may produce the caption ``a cat \textit{holding} a hat.'' Object-level verification confirms both entities are present; attribute-level VQA confirms the hat is white. Neither check, however, addresses the predicate (\textit{holding} versus \textit{wearing}), which requires knowing the physical relationship between the two objects. RelCheck's scene graph layer (RelTR) predicts the triple \textit{(cat, wearing, hat)} with confidence 0.82, and the geometric layer confirms that the hat bounding box overlaps and lies above the cat, consistent with wearing. The corrector revises the caption to ``a cat \textit{wearing} a hat.''

Our main contributions are:
\begin{itemize}
  \item A \textbf{three-layer visual knowledge base} (Claim $\cup$ Scene $\cup$ Geom) that unifies object-level claims, learned relational triples, and deterministic geometric spatial predicates, with explicit reliability ordering.
  \item A \textbf{structured correction pipeline} with a four-level reliability hierarchy, layer-attributed JSON edits, and an edit-distance gate that enforces minimal, evidence-grounded corrections.
  \item An \textbf{ablation study} with four configurations and McNemar significance tests showing that (a) the corrector upgrade alone does not explain the gains, and (b) both relational layers contribute independently.
  \item Experimental evidence that RelCheck consistently improves over a Woodpecker-style baseline across POPE and MME, with gains concentrated exactly on the subtasks that require relational reasoning.
\end{itemize}

\section{Related Work}

\subsection{Hallucinations in MLLMs}

Hallucination in multimodal systems refers to generated text that does not correspond to the input image. The literature organizes multimodal hallucinations into three categories: \textit{object-level} (fabricated or omitted objects), \textit{attribute-level} (incorrect color, count, or state), and \textit{relational} (incorrect spatial or interactive relationships). Rohrbach et al.~\cite{rohrbach2018chair} introduced the CHAIR metric, which quantifies object hallucination in captions by measuring the fraction of mentioned objects absent from ground-truth annotations; this established the problem as measurable, but addresses only object presence.

Zhou et al.~\cite{zhou2023plausible} show that models learn statistical co-occurrence patterns during pretraining, which override visual evidence at test time, a mechanism that applies equally to object- and relation-level predictions. For example, a model that frequently sees ``man'' co-occurring with ``riding horse'' during training will predict the riding predicate at test time even when the image shows a man merely standing beside a horse. This linguistic-prior dominance directly motivates supplying the corrector with structured visual evidence that is independent of the MLLM's learned associations. POPE~\cite{li2023pope} formalizes object hallucination as binary classification, while MME~\cite{fu2023mme} provides structured subtasks covering existence, count, position, and color.

Spatial and relational reasoning is a documented systematic weakness. Kamath et al.~\cite{kamath2023whatsup} demonstrate a 20--30-point accuracy gap between spatial and object queries on controlled images. The Visual Spatial Reasoning benchmark~\cite{liu2023vsr} extends this finding across topological, directional, and distance-based spatial categories. SpatialVLM~\cite{chen2024spatialvlm} addresses spatial reasoning via targeted training, but requires a model-specific synthetic dataset; RelCheck is complementary, operating post-hoc without retraining.

\subsection{Post-Hoc Hallucination Correction}

Post-hoc methods can be grouped into three paradigms. \textit{External verification} pipelines consult independent perception models to ground corrections. Woodpecker~\cite{yin2024woodpecker} is the canonical example: it extracts key concepts from the MLLM's output, generates diagnostic questions, validates them with GroundingDINO~\cite{liu2023groundingdino} and BLIP-2~\cite{li2023blip2}, assembles a visual knowledge base of object-level and attribute-level claims, and rewrites the MLLM's output conditioned on the base. Its limitation is that the knowledge base contains no inter-object relational structure, a gap RelCheck fills by adding scene graph and geometric layers.

\textit{Self-feedback} methods ask the MLLM to critique its own output~\cite{lee2024volcano}, but the model that produced a hallucination is likely to inherit the same visual grounding deficit. LURE~\cite{zhou2024lure} occupies a middle ground, using model uncertainty signals to identify likely-hallucinated outputs before triggering correction; but it requires fine-tuning and cannot be applied to closed-source MLLMs. \textit{Decoding-time intervention} methods such as OPERA~\cite{huang2024opera} penalize tokens with weak visual grounding during generation, but require access to internal attention patterns and cannot correct hallucinations post-hoc.

\subsection{Scene Graph Generation}

Scene graph generation (SGG) predicts structured subject--predicate--object triples from images, where nodes correspond to detected objects and directed edges represent relationships labeled with predicates from a predefined vocabulary. RelTR~\cite{cong2023reltr}, built on DETR~\cite{carion2020detr}, predicts triples end-to-end in a single forward pass using coupled subject and object decoders with a shared predicate branch trained with a Hungarian matching loss. Trained on Visual Genome~\cite{krishna2017visual} ($\sim$108K images, 50 predicate types), it achieves competitive accuracy with significantly lower inference latency than two-stage methods ($\sim$0.3\,s per image on GPU), making it practical for inclusion in a correction pipeline. Its predicate vocabulary covers spatial relations (\textit{on, next to, under}), interaction relations (\textit{holding, riding, wearing}), and possessive relations (\textit{has, part of}).

Despite the maturity of SGG as a field, no prior post-hoc hallucination correction pipeline has incorporated SGG output as a relational evidence layer. Existing pipelines rely on object detectors and VQA models but do not query a dedicated relational model. This is the central novelty of RelCheck.

\section{Methodology}

\subsection{Overview}

\begin{figure*}[t]
\centering
\includegraphics[width=0.96\textwidth, trim=0 30pt 0 0, clip]{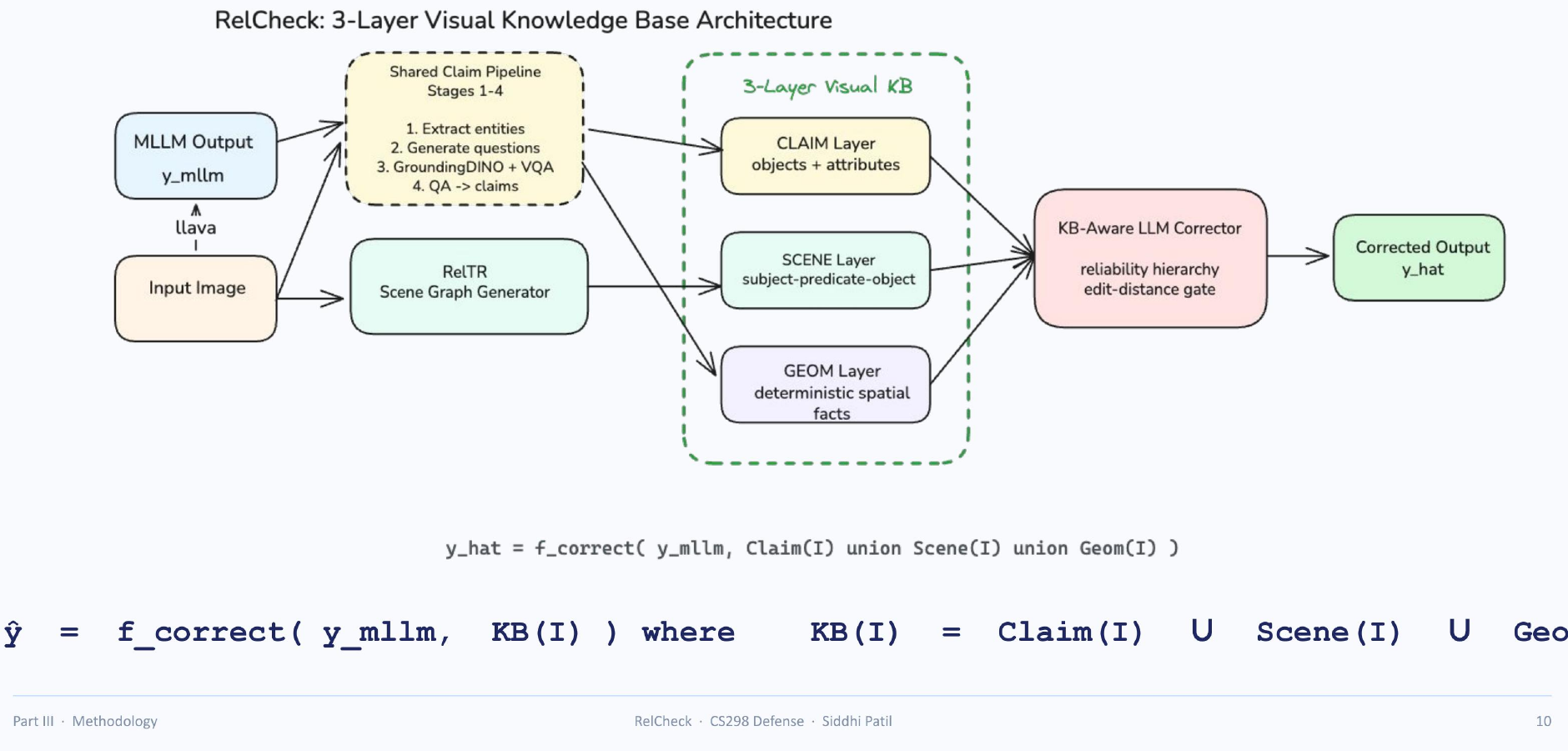}
\caption{RelCheck 3-layer visual knowledge base architecture. Stages 1--4 (shared with Woodpecker) extract entities, generate questions, validate with GroundingDINO and VQA, and produce the Claim layer. RelCheck extends this with two novel layers: the Scene layer from RelTR scene graph generation and the Geom layer from deterministic bounding-box geometry. All three layers feed the KB-aware LLM corrector to produce the grounded output $\hat{y}$.}
\label{fig:pipeline}
\end{figure*}

RelCheck operates post-hoc: it takes an image $I$ and an MLLM-generated response $y_\text{mllm}$ and produces a corrected response $\hat{y}$ grounded in visual evidence. Figure~\ref{fig:pipeline} illustrates the full pipeline. Formally,
\begin{equation}
  \hat{y} = f_\text{correct}\!\left(y_\text{mllm},\, \mathrm{KB}(I)\right),
\end{equation}
where $\mathrm{KB}(I)$ is a multi-layer visual knowledge base derived from $I$ alone, and $f_\text{correct}$ is a language-model corrector. The knowledge base decomposes as
\begin{equation}
  \mathrm{KB}(I) = \mathrm{Claim}(I) \cup \mathrm{Scene}(I) \cup \mathrm{Geom}(I),
\end{equation}
where $\mathrm{Claim}(I)$ provides object-level and attribute-level evidence (identical to Woodpecker's knowledge base), $\mathrm{Scene}(I)$ provides learned relational triples from a scene graph generator, and $\mathrm{Geom}(I)$ provides deterministic spatial predicates from bounding box coordinates.

\subsection{Shared Claim Generation Pipeline (Stages 1--4)}

Both Woodpecker and RelCheck share the same four-stage claim generation pipeline, adapted from~\cite{yin2024woodpecker}.

\textbf{Stage 1 -- Key Concept Extraction.} A language model reads $y_\text{mllm}$ and extracts the main objects and entities mentioned, defining the scope of verification.

\textbf{Stage 2 -- Question Formulation.} Two categories of diagnostic questions are generated: object-level questions (existence, count) and attribute-level questions (color, state, background position).

\textbf{Stage 3 -- Visual Knowledge Validation.} Object-level questions are answered by GroundingDINO~\cite{liu2023groundingdino}, which returns bounding boxes and confidence scores for all extracted concepts. Attribute-level questions are answered by a frontier multimodal language model performing image-conditioned VQA.

\textbf{Stage 4 -- Visual Claim Generation.} Question--answer pairs are converted into declarative claims using a pretrained QA-to-claim T5 model~\cite{chen2022zerofec}, organized into count claims, per-entity specific claims, and scene-level overall claims.

\subsection{Multi-Layer Visual Knowledge Base}

\textbf{Claim Layer.} $\mathrm{Claim}(I)$ is the Woodpecker-style VKB from Stages 1--4. It contains count claims of the form $\mathrm{Obj}(I) = \{(\ell_i, B_i, c_i) : c_i \geq \tau_\text{obj}\}$, where $\ell_i$ is the object label, $B_i$ the normalized bounding box, and $c_i$ the detection confidence, along with per-entity and scene-level attribute claims. This layer is identical between Woodpecker and RelCheck.

\textbf{Scene Graph Layer.} $\mathrm{Scene}(I)$ is produced by RelTR~\cite{cong2023reltr} applied directly to $I$ without any text conditioning:
\begin{equation}
  \mathrm{Scene}(I) = \bigl\{(s_j, p_j, o_j, c_j) : c_j \geq \tau_\text{rel}\bigr\},
\end{equation}
where $s_j$, $o_j$ are subject and object labels, $p_j$ is the predicted predicate from a $\sim$150-class vocabulary trained on Visual Genome~\cite{krishna2017visual}, and $\tau_\text{rel}$ is a confidence threshold.

\begin{figure}[t]
\centering
\includegraphics[width=0.98\columnwidth]{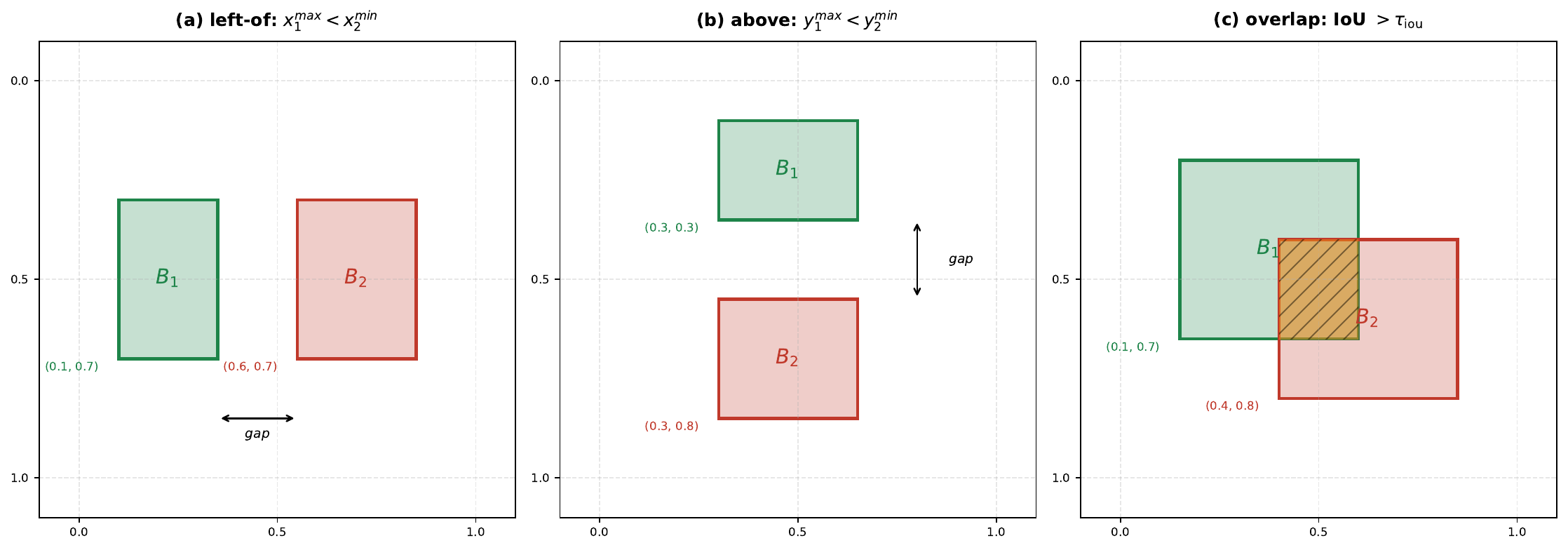}
\caption{Representative geometric predicates from bounding-box coordinates. (a)~left-of: centroid of $B_1$ left of $B_2$ beyond deadzone $T$. (b)~above: centroid of $B_1$ above $B_2$. (c)~on/inside: $B_1 \subseteq B_2$ within 5\% tolerance.}
\label{fig:geom}
\end{figure}

\textbf{Geometric Layer.} $\mathrm{Geom}(I)$ contains deterministic spatial predicates computed from bounding boxes in $\mathrm{Claim}(I)$. For each pair of detected objects $(i, j)$, centroids $(c_{x,i}, c_{y,i})$ and $(c_{x,j}, c_{y,j})$ are compared:
\begin{align}
  \text{left-of}(B_i,B_j) &\equiv c_{x,i} < c_{x,j} - T, \\
  \text{above}(B_i,B_j)   &\equiv c_{y,i} < c_{y,j} - T, \\
  \text{on/inside}(B_i,B_j) &\equiv B_i \subseteq B_j \text{ (5\% tol.)},
\end{align}
where $T$ is a deadzone threshold that prevents spurious predicates for nearly-aligned objects. When neither axis dominates, no predicate is emitted. The geometric layer is entirely model-free and deterministic given the bounding boxes. Figure~\ref{fig:geom} illustrates the three core predicates.

\subsection{Hallucination Correction}

\textbf{Woodpecker Correction.} The claim-layer VKB is concatenated with $y_\text{mllm}$ and a lightweight language model is instructed to ``correct the description based on the given claims.'' The output is unrestricted free-form text.

\textbf{RelCheck Correction.} The full three-layer KB is passed to a more capable language model (gpt-5.4) with \texttt{reasoning\_effort=high}, allocating additional compute for chain-of-thought reasoning before generating the correction. The correction prompt encodes a four-level reliability hierarchy: (1) count claims from GroundingDINO are highly reliable for sparse objects; (2) specific attribute claims from multimodal VQA are trusted given the upstream detection is correct; (3) scene graph triples from RelTR are trusted for the relations they predict, but they may be incomplete (the model returns no evidence for relations it did not find); (4) geometric predicates are reliable for spatially well-separated objects with left/right/above/below relationships, but should be ignored for 3D spatial predicates such as \textit{in front of} and \textit{behind} where 2D bounding boxes are insufficient.

Decision rules specify when each layer's evidence should trigger a correction and when it should be overridden by a higher-confidence layer. For example, if the scene graph contradicts a relational claim but a claim-layer fact supports the original text, the corrector is instructed to trust the claim layer. If only the geometric layer contradicts an \textit{on/under/behind} claim, the corrector is instructed to ignore it.

The corrector returns a structured JSON object listing the corrected caption and each edit, tagged with the KB layer that motivated it (\texttt{CLAIM-Count}, \texttt{CLAIM-Specific}, \texttt{CLAIM-Overall}, \texttt{SCENE}, or \texttt{GEOM}), the specific contradicting evidence, and a confidence level (\texttt{high} or \texttt{medium}). This structured output supports the downstream layer-attribution analysis.

An \textbf{edit-distance gate} (Levenshtein distance $\in [3, 50]$ characters) rejects corrections that are trivially small (no-ops) or wholesale rewrites that risk introducing new hallucinations, enforcing a minimal-edit principle. If a suggested correction falls outside this range, the original response is returned unchanged.

\subsection{Running Example}

\begin{figure*}[t]
\centering
\includegraphics[width=0.96\textwidth]{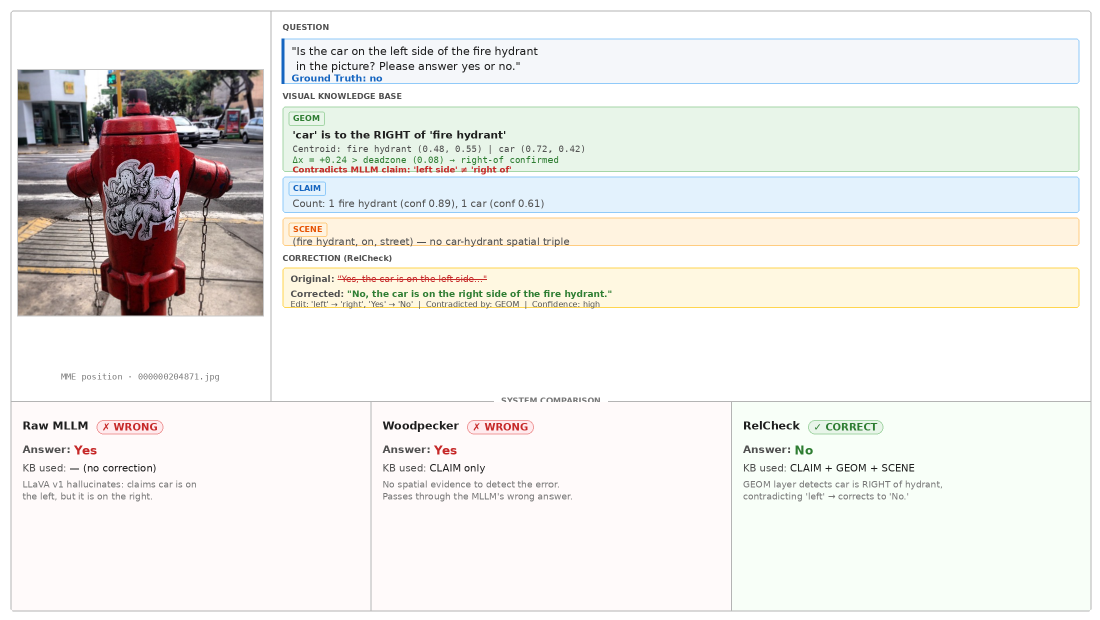}
\caption{Qualitative example: MME position subtask (image 204871). The MLLM responds ``Yes, the car is on the left side of the fire hydrant'' (GT: no). The GEOM layer computes \textit{car right-of fire hydrant} from bounding-box centroids. RelCheck corrects to ``No.'' Woodpecker, lacking geometric evidence, passes the hallucination through unchanged.}
\label{fig:qual_pos}
\end{figure*}

Figure~\ref{fig:qual_pos} traces a concrete example end-to-end. The MLLM hallucinates the spatial relationship (\textit{left} instead of \textit{right}). Woodpecker, lacking any relational layer, passes the hallucination through unchanged. RelCheck's geometric layer computes the deterministic predicate \textit{car right-of fire\_hydrant} from the GroundingDINO bounding box centroids, the corrector identifies the contradiction with \texttt{high} confidence, and replaces ``left'' with ``right,'' flipping the answer from Yes to No.

\subsection{Answer Extraction}

POPE and MME require binary yes/no answers. A zero-shot language model extractor parses the corrected free-form response into a single yes/no token. The same extractor is applied identically to all three systems (RawMLLM, Woodpecker, RelCheck), ensuring that observed differences are attributable to the corrections themselves rather than parsing variation. Answers are cached to avoid redundant API calls for identical (question, response) pairs.

\section{Experimental Setup}

\subsection{Benchmarks}

\textbf{POPE}~\cite{li2023pope} frames object hallucination as binary classification: the model is asked ``Is there a \{object\} in the image?'' across three sampling settings: Random (non-existent objects sampled uniformly), Popular (sampled from the most frequent COCO categories), and Adversarial (sampled from objects that co-occur with the present ones). We evaluate 500 questions per split (1,500 total) on images from COCO val2014~\cite{lin2014coco}. Metrics: accuracy, precision, recall, F1.

\textbf{MME hallucination subset}~\cite{fu2023mme} covers four perception subtasks: \textit{Existence} (object presence/absence), \textit{Count} (specific object counts), \textit{Position} (left/right spatial location), and \textit{Color} (object color). Each subtask has 30 images $\times$ 2 complementary yes/no questions = 60 questions. Metrics: accuracy, accuracy+ (both questions for an image correct), and score = $(\text{accuracy} + \text{accuracy}^+) \times 100 \in [0, 200]$ per subtask.

\subsection{Target MLLM and Implementation}

We use LLaVA~v1~13B~\cite{liu2023llava} with greedy decoding (temperature 0, max 50 tokens) on a single NVIDIA A100 40GB GPU. GroundingDINO is configured with $\tau_\text{obj} = 0.35$, text-matching threshold 0.25. RelTR uses $\tau_\text{rel} = 0.30$. The spatial deadzone is $T = 0.08$.

\subsection{Systems Compared}

Table~\ref{tab:systems} summarizes the three systems compared. All share LLaVA~v1~13B as the target MLLM and the same claim generation pipeline (Stages 1--4), differing only in KB layers and corrector.

\begin{table}[t]
\centering
\caption{System configurations. All share LLaVA v1 13B and Stages 1--4.}
\label{tab:systems}
\setlength{\tabcolsep}{3pt}
\begin{tabular}{lccc}
\toprule
Component & Raw & WP & RC \\
\midrule
GroundingDINO detection   & --  & \checkmark & \checkmark \\
Attribute VQA             & --  & \checkmark & \checkmark \\
QA-to-claim (T5)          & --  & \checkmark & \checkmark \\
Scene graph (RelTR)       & --  & --         & \checkmark \\
Geometric spatial layer   & --  & --         & \checkmark \\
KB layers                 & --  & Claim      & Claim+Scene+Geom \\
Corrector                 & --  & gpt-5.4-mini & gpt-5.4 \\
Reasoning effort          & --  & Standard   & High \\
Edit-distance gate        & --  & --         & \checkmark \\
\bottomrule
\end{tabular}
\end{table}

\textbf{RawMLLM}: Uncorrected LLaVA~v1 output, establishing the hallucination floor.

\textbf{Woodpecker}: Re-implemented Woodpecker~\cite{yin2024woodpecker} using the same claim generation pipeline (Stages 1--4) and GroundingDINO backbone as RelCheck, with a lightweight language model corrector and the original Woodpecker prompt. This isolates the effect of relational grounding.

\textbf{RelCheck}: Full system with all three KB layers, corrected by a more capable language model with high reasoning effort, structured JSON output, and edit-distance gate. Table~\ref{tab:hparams} lists the hyperparameters held constant across all experiments.

\begin{table}[t]
\centering
\caption{RelCheck hyperparameters (fixed across all experiments).}
\label{tab:hparams}
\setlength{\tabcolsep}{4pt}
\begin{tabular}{lll}
\toprule
Component & Parameter & Value \\
\midrule
GroundingDINO & Box confidence $\tau_\text{obj}$ & 0.35 \\
GroundingDINO & Text-matching threshold & 0.25 \\
RelTR         & Triple confidence $\tau_\text{rel}$ & 0.30 \\
Geometry      & Spatial deadzone $T$ & 0.08 \\
Geometry      & Containment tolerance & 5\% \\
Geometry      & Max detections/image & 20 \\
Corrector     & Min edit distance & 3 \\
Corrector     & Max edit distance & 50 \\
\bottomrule
\end{tabular}
\end{table}

\subsection{Implementation Details}

All experiments run on a single NVIDIA A100 40GB GPU. LLaVA~v1~13B is loaded in 8-bit quantization and uses greedy decoding (temperature 0, \texttt{do\_sample=False}) with a maximum of 50 new tokens, yielding fully deterministic outputs. GroundingDINO uses the SwinT-OGC checkpoint with $\tau_\text{obj}=0.35$ and text-matching threshold 0.25. RelTR uses $\tau_\text{rel}=0.30$ with a maximum of 20 detected triples per image.

The LLM components use the OpenAI API: Stages 1--2 (entity extraction, question formulation) and Stage 3b (attribute VQA) use gpt-5.4-mini; the RelCheck corrector uses gpt-5.4 with \texttt{reasoning\_effort=high}; answer extraction uses gpt-5.4-mini text-only. The QA-to-claim model (\texttt{khhuang/zerofec-qa2claim-t5-base}) runs locally. Full evaluation of the 240 MME hallucination samples across all three systems requires approximately 1.5 hours wall-clock; POPE (1,500 samples) scales proportionally. API responses are cached by (prompt, image) hash to avoid redundant calls.

\section{Experimental Results}

\subsection{POPE Evaluation}

\begin{table}[t]
\centering
\caption{POPE results (LLaVA~v1~13B, 500 questions/split). Best per column bolded.}
\label{tab:pope}
\setlength{\tabcolsep}{4pt}
\begin{tabular}{llcccc}
\toprule
Split & System & Acc. & Prec. & Recall & F1 \\
\midrule
\multirow{3}{*}{Random}
 & RawMLLM    & 0.723 & 0.647 & \textbf{0.982} & 0.780 \\
 & Woodpecker & 0.872 & 0.943 & 0.792 & 0.861 \\
 & RelCheck   & \textbf{0.878} & \textbf{0.952} & 0.796 & \textbf{0.867} \\
\midrule
\multirow{3}{*}{Popular}
 & RawMLLM    & 0.678 & 0.611 & \textbf{0.982} & 0.753 \\
 & Woodpecker & 0.838 & 0.872 & 0.792 & 0.830 \\
 & RelCheck   & \textbf{0.852} & \textbf{0.896} & 0.796 & \textbf{0.843} \\
\midrule
\multirow{3}{*}{Adversarial}
 & RawMLLM    & 0.622 & 0.571 & \textbf{0.982} & 0.722 \\
 & Woodpecker & 0.804 & 0.811 & 0.792 & 0.802 \\
 & RelCheck   & \textbf{0.814} & \textbf{0.826} & 0.796 & \textbf{0.811} \\
\bottomrule
\end{tabular}
\end{table}

Table~\ref{tab:pope} presents results across the three POPE splits. The raw LLaVA~v1 baseline exhibits a strong yes-bias: yes-rate is 0.759 / 0.804 / 0.860 across random/popular/adversarial splits, with recall near-ceiling (0.982) while precision falls to 0.571 on the adversarial split. This bias increases from random to adversarial because the adversarial split specifically selects negative objects that co-occur with the present objects in COCO, directly exploiting the model's pretraining co-occurrence prior.

Both correction systems substantially reduce this bias, improving accuracy from the 0.62--0.72 range to 0.80--0.87 (Woodpecker) and 0.81--0.88 (RelCheck). Precision improves most dramatically: from 0.57--0.65 for the raw baseline to 0.81--0.94 for Woodpecker and 0.83--0.95 for RelCheck. This precision improvement reflects the grounding effect of GroundingDINO: when the detector does not find an object, the system correctly answers ``no'' regardless of the MLLM's prior.

RelCheck consistently outperforms Woodpecker by +0.6/+1.4/+1.0 accuracy points across the random/popular/adversarial splits. The modest but consistent gain on an object-existence benchmark suggests that RelTR triples provide auxiliary corroborating evidence: if a scene graph triple involves an object, that object almost certainly exists in the image, giving a second confidence signal independent of GroundingDINO's detection score. This effect is most pronounced on the popular split (+1.4 pp), where frequency-biased objects that may evade detection are more likely to appear as subjects or objects in scene graph triples.

\subsection{MME Hallucination Evaluation}

\begin{figure}[t]
\centering
\includegraphics[width=0.92\columnwidth]{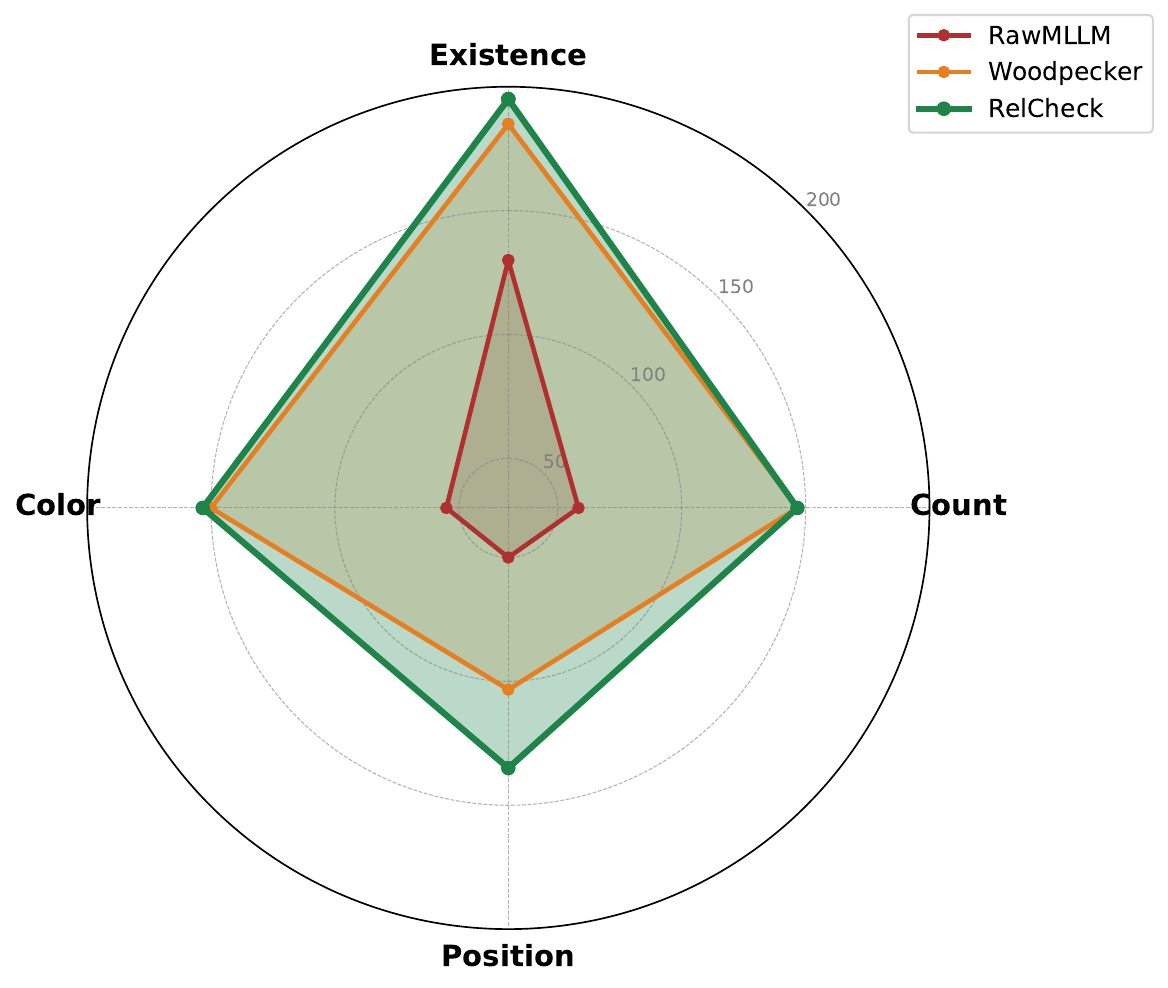}
\caption{Radar chart of MME hallucination scores. RelCheck (green) traces the outer envelope on all four axes; the gap vs.\ Woodpecker is most pronounced on position.}
\label{fig:mme_radar}
\end{figure}

\begin{table}[t]
\centering
\caption{MME hallucination subset results. Score $= (\text{Acc.}+\text{Acc.}^+)\times100$, max 200. Best bolded.}
\label{tab:mme}
\begin{tabular}{llccc}
\toprule
Subtask & System & Acc. & Acc.$^+$ & Score \\
\midrule
\multirow{3}{*}{Existence}
 & RawMLLM    & 0.767 & 0.533 & 130.0 \\
 & Woodpecker & 0.950 & 0.900 & 185.0 \\
 & RelCheck   & \textbf{0.983} & \textbf{0.967} & \textbf{195.0} \\
\midrule
\multirow{3}{*}{Count}
 & RawMLLM    & 0.517 & 0.067 & 58.3 \\
 & Woodpecker & \textbf{0.767} & \textbf{0.700} & \textbf{146.7} \\
 & RelCheck   & \textbf{0.767} & \textbf{0.700} & \textbf{146.7} \\
\midrule
\multirow{3}{*}{Position}
 & RawMLLM    & 0.500 & 0.000 & 50.0 \\
 & Woodpecker & 0.667 & 0.367 & 103.3 \\
 & RelCheck   & \textbf{0.750} & \textbf{0.600} & \textbf{135.0} \\
\midrule
\multirow{3}{*}{Color}
 & RawMLLM    & 0.517 & 0.033 & 55.0 \\
 & Woodpecker & \textbf{0.833} & 0.667 & 150.0 \\
 & RelCheck   & \textbf{0.833} & \textbf{0.700} & \textbf{153.3} \\
\midrule
\multicolumn{2}{l}{\textbf{Total}} & & & \\
 & RawMLLM    & -- & -- & 293.3 \\
 & Woodpecker & -- & -- & 585.0 \\
 & RelCheck   & -- & -- & \textbf{630.0} \\
\bottomrule
\end{tabular}
\end{table}

Table~\ref{tab:mme} presents the detailed MME results. RelCheck achieves a total score of 630.0 versus 585.0 for Woodpecker, a 45-point improvement over an already strong baseline and a 336.7-point improvement over the raw MLLM.

\textbf{Position (+31.7 over Woodpecker).} This subtask yields the headline result. The raw MLLM scores 50.0 (chance on a balanced binary task) with accuracy+ of 0.000, meaning it cannot simultaneously answer both the positive and negative positional questions for any image. Woodpecker improves to 103.3 by narrowing candidate positions through object detection. RelCheck reaches 135.0, with accuracy+ rising from 0.367 to 0.600 (+63.5\% relative). The geometric layer's deterministic left/right predicates provide the exact evidence needed to confirm or deny left/right position claims (the only relationship type the MME position subtask tests), provided GroundingDINO has detected the relevant objects.

\textbf{Existence (+10.0 over Woodpecker).} RelCheck attains 195.0 versus 185.0 for Woodpecker, driven by accuracy+ of 0.967 vs.\ 0.900. RelTR triples provide corroborating evidence: an object appearing in a scene graph triple is very likely present, giving a second signal beyond GroundingDINO confidence alone.

\textbf{Count (tied, 146.7).} RelCheck and Woodpecker achieve identical scores. Object counting relies exclusively on GroundingDINO instance detection, which is shared between the two systems. Neither relational triples nor geometric predicates carry counting information. This equality serves as a clean control: it demonstrates that the other improvements are not artifacts of system-level differences unrelated to relational grounding.

\textbf{Color (+3.3 over Woodpecker).} Both systems achieve the same accuracy (0.833), but RelCheck's accuracy+ is 0.700 vs.\ 0.667. A small gain arises when a RelTR triple (for example, \textit{(red car, parked on, street)}) provides disambiguating color context that supplements the VQA answer. Interestingly, the ablation (Table~\ref{tab:ablation}) shows that SCENE alone causes a minor regression (155.0 $\to$ 153.3) before GEOM is added, suggesting that scene graph predicates occasionally distract the corrector on color questions when they introduce irrelevant relation evidence.

\subsection{Qualitative Analysis}

\begin{figure*}[t]
\centering
\includegraphics[width=0.96\textwidth]{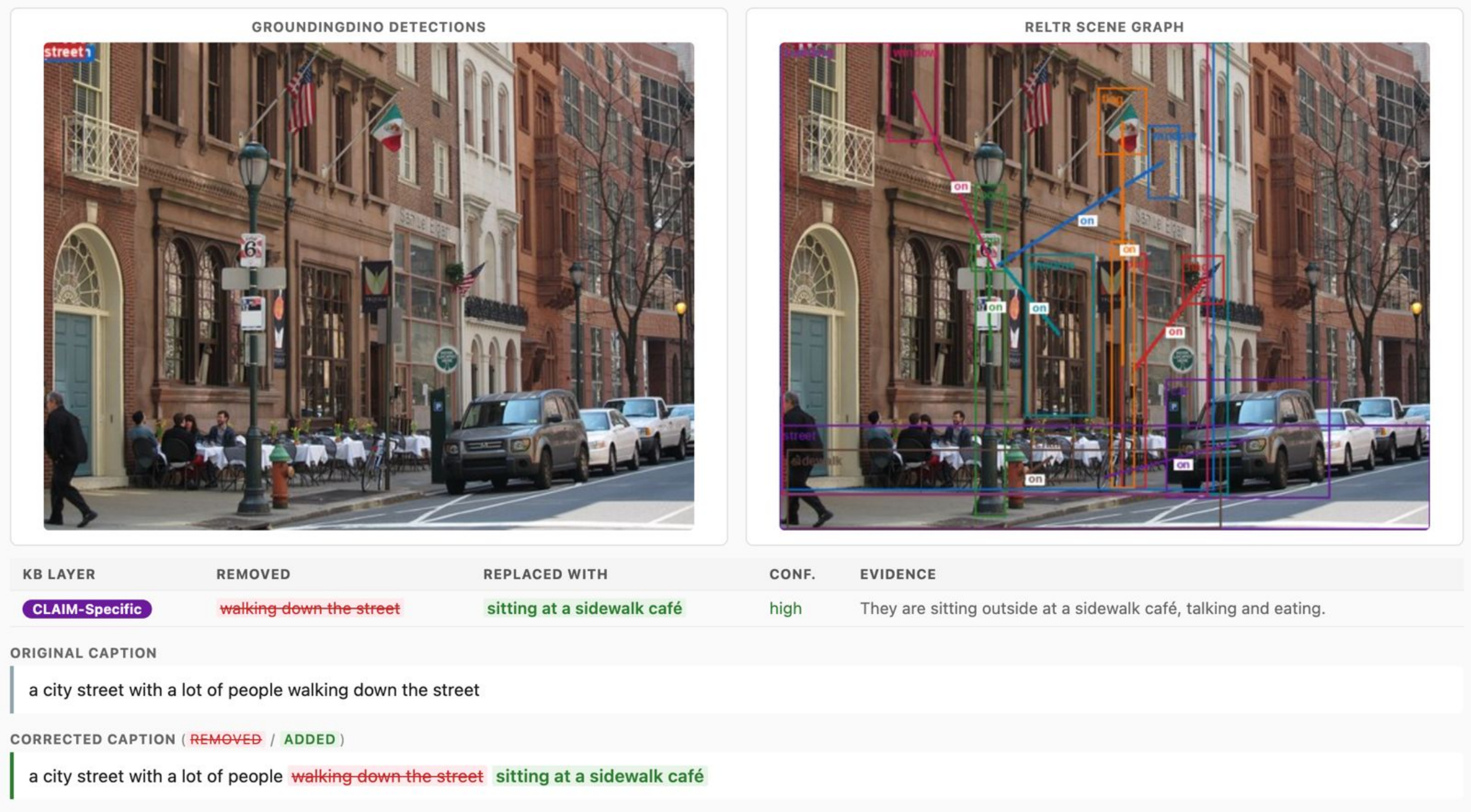}
\caption{Action-relation correction. The MLLM describes ``multiple people walking down the street,'' but the Scene layer (RelTR) detects \textit{sitting} relations and the Claim layer confirms the setting as a caf\'e sidewalk via VQA. RelCheck corrects to ``multiple people sitting at a caf\'e's sidewalk,'' replacing the hallucinated action predicate with the grounded one.}
\label{fig:qual_act}
\end{figure*}

Figure~\ref{fig:qual_pos} already shows the position correction example (image 204871). A second instructive case (Figure~\ref{fig:qual_act}) involves action-relation correction. For a street scene image, LLaVA~v1 generates ``multiple people walking down the street.'' The scene graph layer (RelTR) produces the triples \textit{(people, sitting, sidewalk)} with confidence 0.84 and \textit{(people, at, caf\'e\_table)} with confidence 0.79. The claim layer independently confirms the setting as a caf\'e sidewalk through VQA (``They are sitting outside at a sidewalk caf\'e, talking and eating.''). RelCheck's corrector identifies ``walking down the street'' as contradicted by CLAIM-Specific evidence and replaces it with ``sitting at a caf\'e's sidewalk,'' correcting both the action (\textit{walking}$\to$\textit{sitting}) and the scene context (\textit{street}$\to$\textit{caf\'e sidewalk}).

This example illustrates that the scene graph layer captures interaction-type relational hallucinations that neither GroundingDINO nor VQA alone would detect.

\subsection{Ablation Study}

Table~\ref{tab:ablation} presents a four-configuration ablation isolating the contribution of each KB layer from the corrector upgrade. \textit{ClaimOnly} uses the upgraded corrector (gpt-5.4, reasoning\_effort=high) with the claim layer only. \textit{+GEOM} adds the geometric layer. \textit{+GEOM+SCENE} (RelCheckFull) adds both relational layers.

\begin{table}[t]
\centering
\caption{Ablation on MME hallucination subset (scores $= (\text{Acc}+\text{Acc}^+)\times 100$).}
\label{tab:ablation}
\setlength{\tabcolsep}{3.5pt}
\begin{tabular}{lcccc}
\toprule
Configuration & Exist. & Count & Pos. & Total \\
\midrule
Woodpecker         & 185.0 & 146.7 & 103.3 & 585.0 \\
ClaimOnly          & 195.0 & 146.7 & 105.0 & 580.0 \\
+GEOM              & 195.0 & 146.7 & 121.7 & 618.3 \\
+GEOM+SCENE (Full) & 195.0 & 146.7 & \textbf{135.0} & \textbf{630.0} \\
\bottomrule
\end{tabular}
\end{table}

Three findings emerge. (1) ClaimOnly $\approx$ Woodpecker (580 vs.\ 585, McNemar $p=0.45$): the corrector upgrade alone does not account for the gain. (2) The position score improves additively (+1.7 from the corrector, +16.7 from GEOM, +13.3 from SCENE), confirming that both layers contribute independently. (3) The overall transition from ClaimOnly to RelCheckFull is statistically significant (McNemar $p=0.025$), establishing that the relational evidence layers are responsible for the improvement.

Table~\ref{tab:layerattr} shows how frequently each KB layer produces non-empty evidence per subtask. GEOM fires on 87\% of Position samples (52/60) but rarely elsewhere, directly linking its activation pattern to the position gain. This causal specificity (large activation exactly where the gain is largest, near-zero elsewhere) supports the interpretation that the geometric layer drives position improvement rather than a general corrector effect.

\begin{table}[t]
\centering
\caption{Layer activation: images (out of 60) for which each KB layer produces non-empty evidence per MME subtask.}
\label{tab:layerattr}
\begin{tabular}{lcccc}
\toprule
Subtask & CLAIM & GEOM & SCENE & Gain \\
\midrule
Existence & 58 & \phantom{0}1 & 58 & $+10.0$ \\
Count     & 60 & \phantom{0}0 & 60 & $\phantom{+}0.0$ \\
Position  & 60 & \textbf{52} & 58 & $+31.7$ \\
Color     & 59 & \phantom{0}4 & 60 & $+3.3$ \\
\bottomrule
\end{tabular}
\end{table}

\subsection{Disagreement Analysis}

Of 240 MME samples, RelCheck and Woodpecker disagree on 35 (RelCheck wins 21, Woodpecker 14). Position dominates: RelCheck wins 13 versus 8 for Woodpecker. Count and color cancel out (2--2, 4--4), confirming relational layers provide no systematic harm where spatial evidence is absent. In the 8 cases where Woodpecker wins, the GEOM layer correctly identifies the spatial predicate but the corrector fails to apply it due to conflicting signals, indicating room for improved conflict-resolution prompting.

\section{Conclusion}

We presented RelCheck, a training-free post-hoc hallucination correction pipeline that augments object-level visual grounding with explicit relational evidence. By adding a scene graph layer (RelTR) and a deterministic geometric layer to the Woodpecker-style claim layer, RelCheck constructs a three-layer visual knowledge base that captures the spatial and interactional relationships between objects that prior methods miss.

On the MME hallucination benchmark, RelCheck outperforms a Woodpecker-style baseline by 45 points (total score 630.0 vs.\ 585.0), with 31.7 of those points concentrated on the position subtask, precisely the subtask most sensitive to relational grounding. Accuracy+ on position improves from 0.367 to 0.600, reflecting the ability to consistently answer both the positive and negative variants of a spatial question. RelCheck also consistently outperforms Woodpecker across all three POPE splits, and the tied scores on the count subtask serve as a built-in control confirming that gains are attributable to relational evidence rather than confounding system differences.

A four-configuration ablation confirms that the corrector upgrade alone does not explain the gains (ClaimOnly $\approx$ Woodpecker, $p=0.45$), while the transition to the full three-layer system is statistically significant ($p=0.025$). The position improvement decomposes cleanly: +1.7 from the corrector, +16.7 from GEOM, and +13.3 from SCENE.

The primary limitation is the pipeline's dependence on GroundingDINO's open-vocabulary detection: entities not detected yield no GEOM or VQA evidence, allowing some spatial hallucinations to pass through uncorrected. Additionally, the 2D geometric layer cannot evaluate 3D relations such as \textit{in front of} or \textit{behind}. Current experiments cover only LLaVA~v1~13B; future work should evaluate on InstructBLIP, LLaVA-1.5, and LLaVA-NeXT, incorporate a monocular depth estimator as a fourth KB layer for 3D predicates, and replace API-based components with open-source alternatives to reduce cost and improve reproducibility.

These results confirm that relational hallucinations are a distinct and correctable failure mode in MLLMs, and that training-free, plug-and-play relational grounding is a practical path toward addressing them.

\textbf{Broader Impact.} Accurate spatial grounding in MLLMs has direct downstream value across science and engineering: in medical imaging, lesion positions and anatomical relationships must be described correctly; in robotics, scene understanding depends on accurate interaction predicates; in scientific literature mining, relational claims between entities must be faithfully extracted. RelCheck's training-free design allows it to be layered on top of any existing MLLM without retraining, enabling immediate deployment in these domains. More broadly, augmenting learned model outputs with structured, deterministic geometric evidence is a reusable pattern for grounding AI in verifiable physical facts, with impact wherever spatial accuracy of AI-generated descriptions is quality- or safety-critical.

\bibliographystyle{IEEEtran}

\end{document}